\documentclass[10pt,twocolumn,letterpaper]{article}

\usepackage{cvpr}
\usepackage{times}
\usepackage{epsfig}
\usepackage{graphicx}
\usepackage{amsmath}
\usepackage{amssymb}
\usepackage{multirow}
\usepackage[table]{xcolor}
\usepackage{hhline}
\usepackage{float}
\usepackage{subcaption}
\usepackage[outercaption]{sidecap}    
\usepackage{fancyhdr}

\usepackage[pagebackref=true,breaklinks=true,letterpaper=true,colorlinks,bookmarks=false]{hyperref}

\cvprfinalcopy 

\def\x{\textbf{x} }
\def\p{\textbf{p} }

\font\cvprtenhv  = phvb at 8pt

\ifcvprfinal\fi

\begin{document}

\title{An Empirical Evaluation of Visual Question Answering for Novel Objects
} 

\author{Santhosh K. Ramakrishnan$^{1, 2}$
\quad
Ambar Pal$^{1}$
\quad
Gaurav Sharma$^{1}$
\quad
Anurag Mittal$^{2}$
\vspace{0.4em}\\
$^{1}$IIT Kanpur\thanks{The project started when Santhosh Ramakrishnan
and Ambar Pal were summer interns at IIT Kanpur. Ambar Pal is a student at IIIT Delhi.
{\tt \footnotesize ambar14012@iiitd.ac.in, grv@cse.iitk.ac.in}
} 
\quad \quad \quad 
$^2$IIT Madras\thanks{
{\tt \footnotesize \{ee12b101@ee, amittal@cse\}.iitm.ac.in}}
}
\maketitle

\definecolor{Gray}{gray}{0.95}
\def\etal{et al\onedot}
\def\etc{etc\onedot}
\def\ie{i.e\onedot}
\def\eg{e.g\onedot}
\def\cf{cf\onedot}
\def\vs{vs\onedot}
\def\grad{\nabla}
\def\b{\textbf{b}}
\def\v{\textbf{v}}
\def\a{\boldsymbol{\alpha}}
\def\sign{\textrm{sign}}
\def\pd{\partial}
\def\T{\mathcal{T}}
\def\R{\mathbb{R}}
\def\Reg{\mathcal{R}}
\def\X{\mathcal{X}}
\def\I{\mathcal{I}}
\def\F{\mathcal{F}}
\def\Obj{\mathcal{O}}
\def\V{\mathcal{V}}
\def\w{\textbf{w}}
\def\Dt{{D_2}}
\def\c{\textbf{c}}
\def\1{\textbf{1}}
\def\x{\textbf{x}}
\def\c{\textbf{c}}
\def\s{\textbf{s}}
\def\d{\boldsymbol{\delta}}
\def\y{\textbf{y}}
\def\l{\textbf{l}}
\def\TODO{\textcolor{red}{TODO} }

\begin{abstract}
We study the problem of answering questions about images in the harder setting, where the test
questions and corresponding images contain novel objects, which were not queried about in the
training data. Such setting is inevitable in real world---owing to the heavy tailed distribution of
the visual categories, there would be some objects which would not be annotated in the train set. We
show that the performance of two popular existing methods drop significantly (up to $28\%$) when
evaluated on novel objects \cf known objects. We propose methods which use large existing external
corpora of (i) unlabeled text, \ie books, and (ii) images tagged with classes, to achieve novel
object based visual question answering. We do systematic empirical studies, for both an oracle case
where the novel objects are known textually, as well as a fully automatic case without any explicit
knowledge of the novel objects, but with the minimal assumption that the novel objects are
semantically related to the existing objects in training. The proposed methods for novel object
based visual question answering are modular and can potentially be used with many visual question
answering architectures. We show consistent improvements with the two popular architectures
and give qualitative analysis of the cases where the model does well and of those where it
fails to bring improvements.
\vspace{-1em}
\end{abstract}
\section{Introduction}
Humans seamlessly combine multiple modalities of stimulus, \eg audio, vision, language, touch,
smell, to make decisions. Hence, as a next step for artificial intelligence, tasks involving such
multiple modalities, in particular language and vision, have attracted substantial attention
recently. Visual question answering (VQA), \ie the task of answering a question about an image, has
been recently introduced in a supervised learning setting \cite{malinowski15iccv, AntolICCV2015}. In
the currently studied setup, like in other supervised learning settings, the objects in the training
data and the test data overlap almost completely, \ie all the objects that appear during testing
have been seen annotated in the training. This setting is limited as this requires having training
data for all possible objects in the world---this is an impractical requirement owing to the heavy
tailed distribution of the visual categories. There are many objects, on the tail of the
distribution, which are rare and annotations for them might not be available. While humans are
easily able to generalize to novel objects, \eg make predictions and answer questions about a wolf,
when only a cat and/or a dog were seen during training, automatic methods struggle to do so. In the
general supervised classification, such a setting has been studied as \emph{zero shot learning}
\cite{LEB08}, and has been applied for image recognition as well \cite{HEEY15, LampertPAMI2013,
XianCVPR2016, YA10}. While the zero shot setup works with the constraint that the test classes or
objects were never seen during training, it also assumes some form of auxiliary information to
connect the novel test classes with the seen train classes. Such information could be in the form of
manually specified attributes \cite{HEEY15, LampertPAMI2013, YA10} or in the form of relations
captured between the classes with learnt distributed embeddings like, \texttt{Word2Vec}
\cite{MikolovNIPS2013} or \texttt{GloVe} \cite{PenningtonEMNLP2014}, of the words from an
unannotated text corpus \cite{XianCVPR2016}. In the present paper, we are interested in a similar
setting, but for the more unconstrained and challenging task of answering questions about novel
objects present in an image. Such a setting, while being natural, has not been studied so far, to
the best of our knowledge.

\begin{figure}
\centering
\includegraphics[width=0.95\linewidth, trim=0 280 250 0, clip]{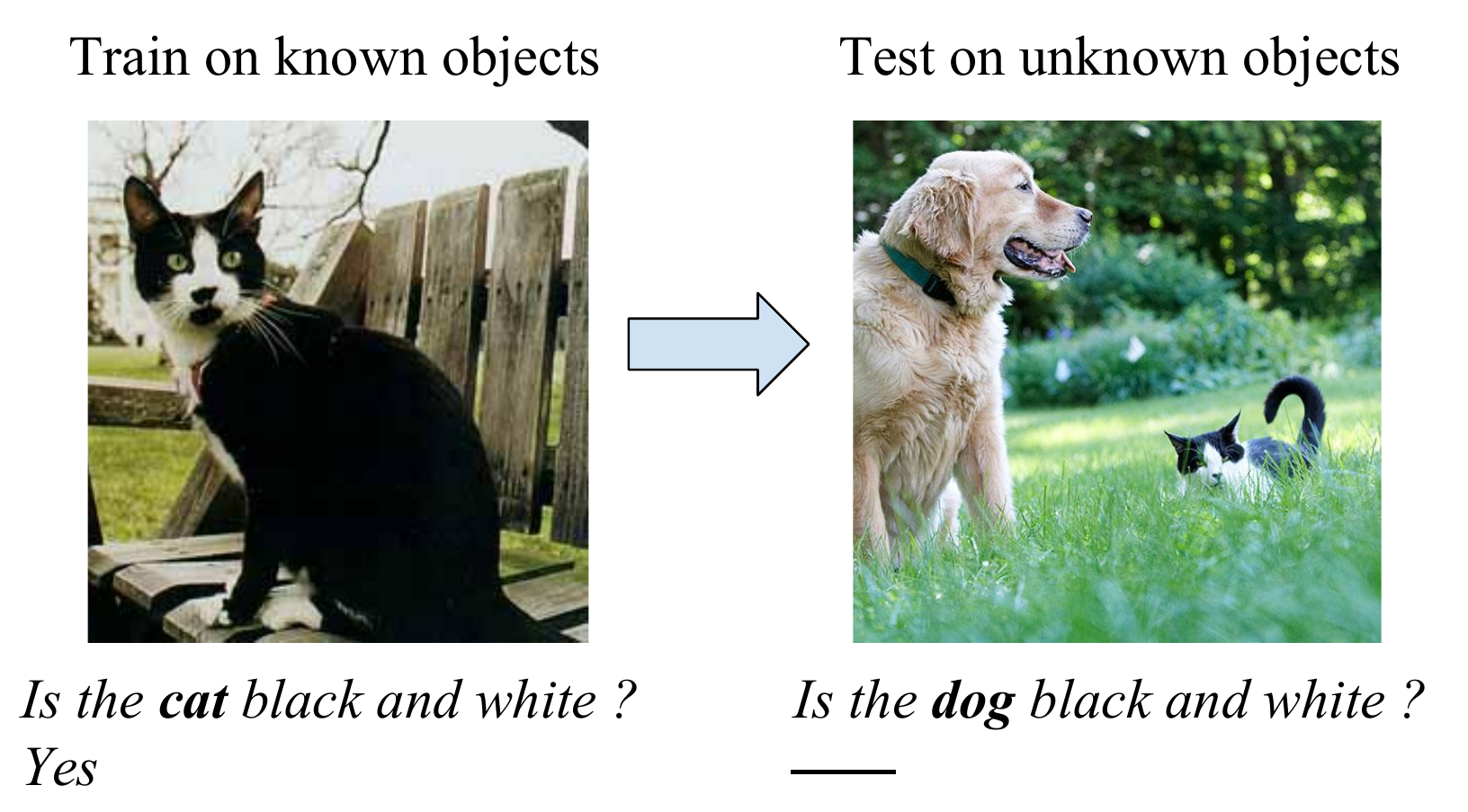}
\vspace{-0.8em}
\caption{We are interested in answering questions about images containing objects not seen at
training.
}
\vspace{-1em}
\label{fig:baseline1}
\end{figure}

We start studying the problem by first proposing a novel split (\S \ref{labelSplit}), into train and
test sets, of the large-scale public dataset for VQA recently proposed by
Antol \etal \cite{AntolICCV2015}. Our split ensures that the novel objects in the test set are never
seen in the train set; we select the novel objects and put all the questions that contain those
objects into the test set along with all the questions whose answers contain them as well. This
means that the train set does not contain any question which either (i) makes a query about the
novel objects, or (ii) queries about some aspect of the image which indicates any of the novel
objects, \ie has any possible answer mentioning the novel object. Hence, the split is strong 
as any information about the novel object is missing from the train set.

We then take two deep neural network based architectures which have shown good performance on tasks
based on language and vision combined \cite{Lu2015, ren2015exploring}. We benchmark them on the new 
split for novel object VQA and compare the performances on the known object setting. As
expected, we find that the performances drop significantly (up to $28\%$) when there are novel
objects in the test set. We then propose two methods based on deep recurrent neural network based
multimodal autoencoder, which exploit large existing auxiliary datasets of text and images, to answer
questions about novel objects, with the two architectures studied. The proposed frameworks are
modular and can be used with many neural networks based VQA systems.  We show that the proposed
methods improve the performance of the system, equally when (i) an oracle is assumed, that gives the
novel test objects and (ii) when the minimal assumption is made that the novel test objects are
semantically related, quantified by their similarity in distributed \texttt{Word2Vec} embedding
space \cite{MikolovNIPS2013}, to the train objects. We extensively study multiple configurations
quantitatively and also analyse the results qualitatively to show the usefulness of the proposed
method in this novel setting. 

\section{Related works}
Image based question answering was introduced by Malinowski and Fritz \cite{malinowski2014towards}
as the \emph{Visual Turing Test}. With the large scale dataset, introduced by Antol \etal
\cite{AntolICCV2015}, recently there has been a lot or interest in the problem. The survey by Wu
\etal \cite{wu2016visual} categorizes the methods for VQA into three categories. First, the joint
embedding based methods, which bring the visual and textual vectors into a common space and then
predict the answer~\cite{FukuiPYRDR16, GaoMZHWX15, malinowski15iccv, Noh_2016_CVPR, ren2015image,
AndreasRDK15, AndreasRDK16}, second, attention based systems which focus on the relevant spatial
regions in the images which support the question \cite{ChenWCGXN15, LuYBP16, shih2016wtl, XuS15a,
yang2015stacked, ZhuGBF15} and finally, third, which are based on networks with explicit memory
mechanisms \cite{KumarISBEPOGS15, XiongMS16}.

Malinowski \etal~\cite{malinowski15iccv} and Gao \etal~\cite{GaoMZHWX15} encode the question and
image using an LSTM and use a decoding LSTM to generate the answers.  Ren \etal~\cite{ren2015image}
predict a word answer using a multi-class classification over a pre-defined vocabulary of single
word answers. Fukui \etal~\cite{FukuiPYRDR16} propose a multimodal bilinear pooling, using Fourier
space computations for efficiency.

Zhu \etal~\cite{ZhuGBF15} augment the LSTM with spatial attention, by learning weights over the
convolutional features. Similarly, Chen \etal~\cite{ChenWCGXN15} generate a question-guided
attention map using convolution with a learnt kernel.  Yang \etal~\cite{yang2015stacked} use stacked
attention networks that iterate to estimate the answer. Xu \etal~\cite{XuS15a} propose a multi-hop
image attention scheme, where the two types of hops are guided by word-based and question-based
attention.  Shih \etal~\cite{shih2016wtl} use region proposals to find relevant regions in the image
\wrt the question and potential answer pairs.  Lu \etal~\cite{LuYBP16} propose a hierarchical
co-attention model where both image and question steer the attention over parts of each other. 

Dynamic Memory Networks of Kumar \etal~\cite{KumarISBEPOGS15} and their variants \cite{WestonCB14,
SukhbaatarSWF15, BordesUCW15}, have been recently adapted and applied to VQA by Xiong \etal
\cite{XiongMS16}. They use an explicit memory to read and write depending on
the input question, allowing them to understand the questions better. 

Methods which use auxilary image or text datasets or other sources of knowledge have also been
proposed.  Wang \etal~\cite{WangWSHD15, WangWSHD16} propose methods which use knowledge bases for
VQA.  Wu \etal~\cite{WuWSHD15} predict semantic attributes in the image and exploit external
knowledge bases to query for related knowledge, to better understand the question.

Similar in spirit to the current work, zero shot learning, \ie when the set of test classes is
disjoint from the set of train classes, has been well studied in the literature \cite{HEEY15,
LampertPAMI2013, LEB08, YA10}. Zero shot learning aims to predict novel object categories without
any visual training examples but with auxilary relations between the known and unknown objects, \eg
in the form of common attributes. Lampert \etal~\cite{LampertPAMI2013} proposed to use attributes
for zero shot image classification while more recent work by Xian \etal~\cite{XianCVPR2016} showed
that it could be achieved using embeddings learnt from unsupervised text data. Most of the current
state-of-the-art methods for zero shot classification use an embedding based approach where the images
and classes (the word for the class, \eg `dog', `cat') are embedded into respective spaces and a
bilinear compatibility function is learnt to associate them \cite{frome2013devise, XianCVPR2016}. 

Our work is also related to the recent works on autoencoders for vector sequences based on recurrent
neural networks (RNN). Such autoencoders have been recently used in text processing
\cite{li2015hierarchical, dai2015semi} as well for doing semi-supervised learning and fine tuning of
RNN based language models. 

\section{Approach}
\begin{SCfigure*}
\centering
\includegraphics[width=0.8\columnwidth, trim=70 300 180 130, clip]{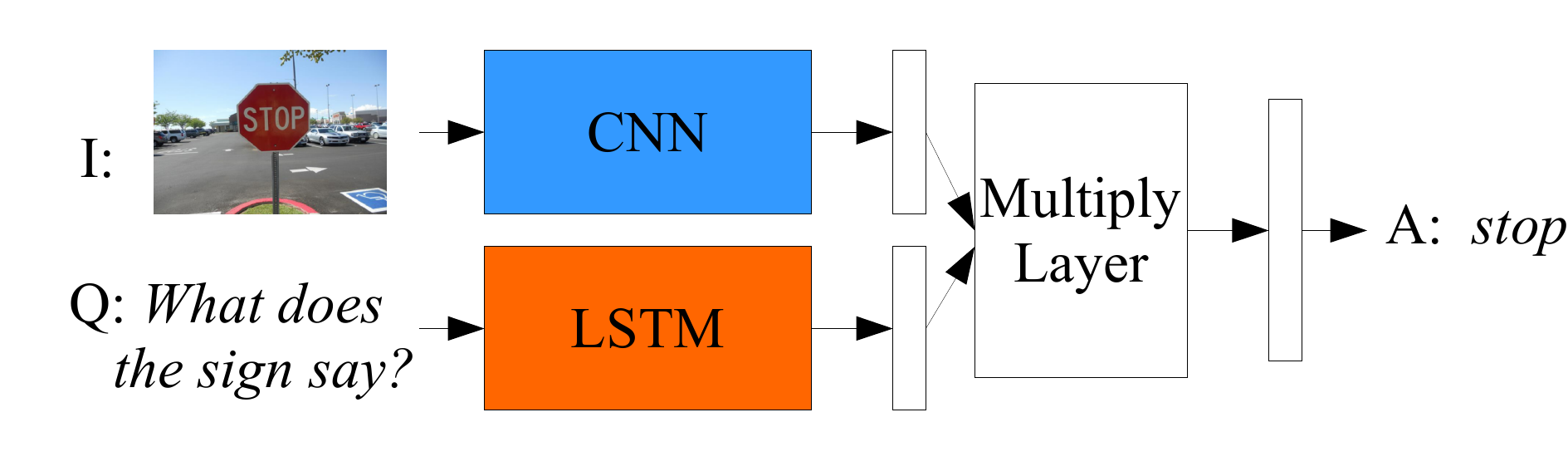}
\hspace{0.5em}
\includegraphics[width=0.8\columnwidth, trim=0 350 280 65, clip]{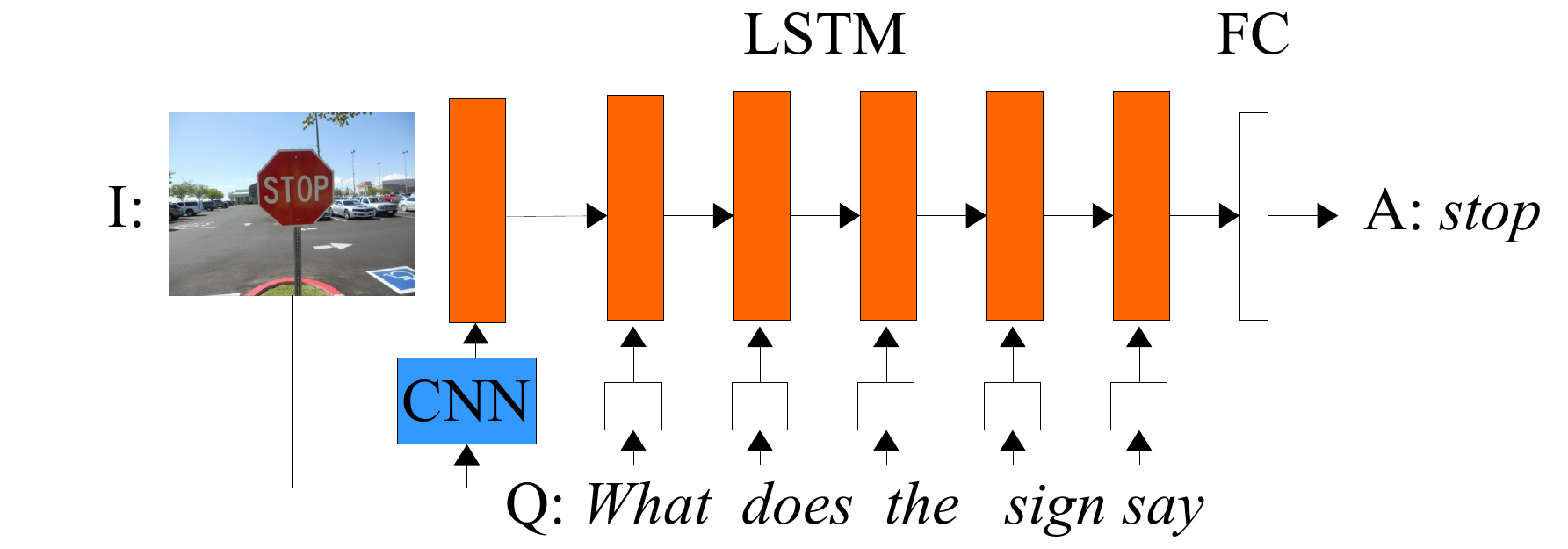}
\vspace{-1em}
\caption{The two Visual Question Answering (VQA) architectures used.}
\label{figArchs}
\end{SCfigure*}

We are interested in extending the VQA models to better answer questions about novel objects by
being aware of them both textually and visually. Towards that end, we start with two existing
architectures, for VQA, and expose them to extra information, from auxiliary datasets of text and
images, in a carefully designed manner. This allows them to be able to answer questions about novel
objects that are not present in the VQA training data. 
We consider two successful deep neural network based architectures, illustrated in
Figure~\ref{figArchs}, whose variants have been used in recent literature~\cite{Lu2015,
ren2015exploring}. We first describe the base architectures and then give the proposed
training and architectural extensions for novel object induction.

\subsection{Base Architectures}
\noindent
\textbf{Architecture 1.} 
The first architecture, shown in Figure~\ref{figArchs} (left),
proposed by Lu~\etal~\cite{Lu2015}, uses a Long Short Term Memory (LSTM) based recurrent neural
network, to encode the question as $\x_Q \in \R^{d_Q}$, and a Convolutional Neural Network (CNN) to
encode the image as $\x_I \in \R^{d_I}$. The two encoded representations are then projected to a
common multimodal space with projection matrices $W_Q \in \R^{d \times d_Q}$ and $W_I \in \R^{d
\times d_I}$ respectively. The projected vectors are then multiplied element-wise to obtain the
joint multimodal representation of the question and the image. This representation is then, in turn,
projected to the answer space using a fully connected layer to obtain probabilities over the set of
possible answers,
\vspace{-0.7em}
\begin{equation}
    \p_{QI} =  W_{QI} \left( \tanh(W_Q \x_Q) \odot \tanh(W_I \x_I) \right).\vspace{-0.5em}
\end{equation}
Here, $\p_{QI}$ is the unnormalized probability distribution over the set of all possible answers, 
given the image, question pair \ie the model treats the VQA task as a multimodal signal
classification task. The answer with the maximum probability is then taken as the predicted answer.

\vspace{0.4em} 
\textbf{Architecture 2.} 
The second architecture, shown in Figure~\ref{figArchs} (right), proposed by Ren
\etal~\cite{ren2015exploring}, borrows ideas from image captioning literature. It
treats the image as the first word of the question, by projecting the image feature vector
$\x_I$ to the word embedding space with a learnt projection matrix $W_e$. Following the image
first, the question words are then passed one at a time to the LSTM. The hidden state vector of
the LSTM after the last time step, which now becomes the joint embedding of the question and the
image, is then projected to the answer space to obtain the probabilities over the set of answers,
similar to Architecture 1 above. 
\vspace{-0.2em}
\subsection{Inducing novel objects using auxiliary datasets}
\label{secNovObj}
\begin{SCfigure*}
\centering
\includegraphics[width=0.88\columnwidth, trim=20 310 110 130, clip]{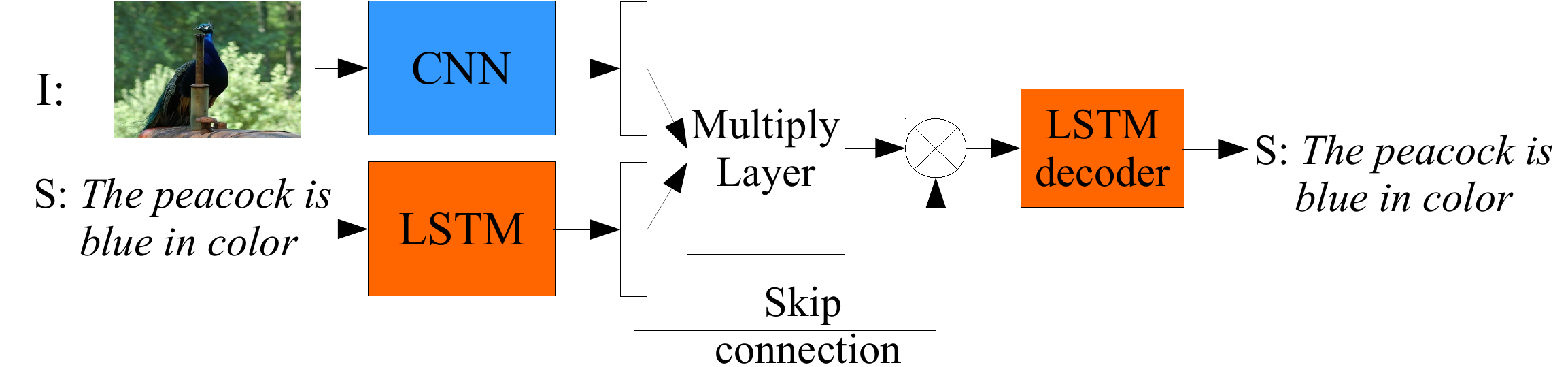}
\includegraphics[width=0.75\columnwidth, trim=0 375 340 60, clip]{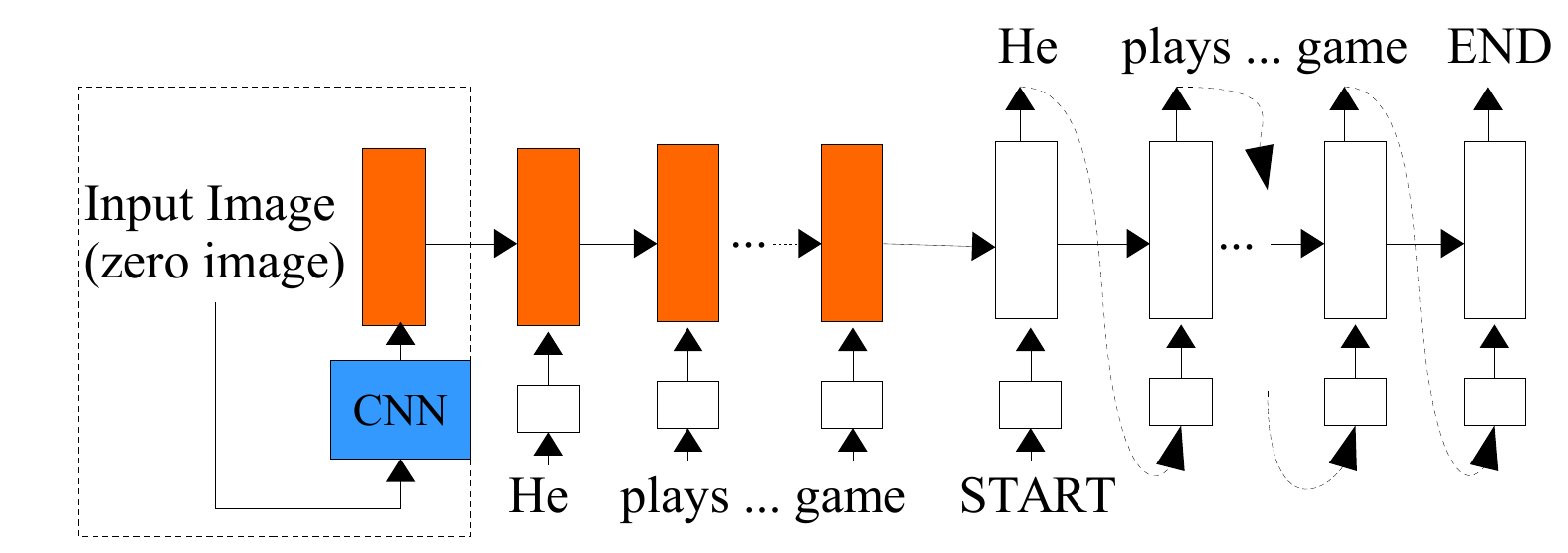}
\vspace{0.7em}
\caption{Autoencoders used to pre-train the respective VQA networks' encoders.}
\vspace{-1.5em}
\label{figAE}
\end{SCfigure*}

Given the above two architectures, we now explain how we introduce novel objects using auxiliary
datasets. We experiment with two different settings, first,  when the novel words are known
textually, and, second, when the novel words are not known. The former is similar to the zero-shot
classification \cite{LampertPAMI2013} setting where the unknown classes are never seen visually at
training but are known textually. In the latter, we make the assumption that the novel words are
semantically close to the known words; where we use the vector similarity of the words in a
standard distributed word embedding space, \eg \texttt{word2vec} \cite{mikolov2013distributed}. 
Given the novel words from the two settings, to make the system aware of novel concepts, we have two
sources of auxiliary information. We could use large amount of text data, \eg from Wikipedia or
books, as well as image data from large datasets such as ImageNet \cite{ILSVRC15}.  We now describe
the different  ways in which we propose to exploit such auxiliary datasets for making the above
described VQA systems aware of novel objects.
\vspace{0.4em} \\
\textbf{Auxiliary text data only.} 
In the first method we propose to use only auxiliary text data for improving VQA performance for
novel objects. In most of the VQA architectures, the question encoding is done with a recurrent
neural network such as the LSTM network. When large amount of text data is available, which contains
both the known and novel objects and the relations between them (as could be described textually),
we hypothesise that pre-training question encoder on the auxiliary dataset could be beneficial.
To pre-train the question encoder, we use an LSTM based sequence autoencoder (AE), \eg
\cite{dai2015semi, li2015hierarchical}.  The AE is pre-trained on a large external text dataset, \eg
BookCorpus \cite{moviebook}. Figure~\ref{figAE} illustrates the AEs (with the
dashed block absent, we explain it more below).

However, this is not a straightforward pre-training as the text vocabulary needs to be expanded to
contain the novel words, so that the VQA system is aware of them and does not just see them as
\texttt{UNK} (special token for all words not in vocabulary). It could be argued that
pre-training with only the current vocabulary may improve the encoder in general and might help the
VQA system---we test this system as well in the experiments. To do such vocabulary expansion is
non-trivial; we could use a vocabulary from the external corpus\footnote{All words with frequency
above a threshold in the whole dataset}, but such a vocabulary turns out to be very large and can
degrade the VQA performance. Thus, we evaluate two ways to construct the vocabulary. 
\vspace{0.3em} \\
\textbf{Oracle setting.} 
First, we assume an oracle setting where we know (textually) the novel words that will
appear---this is similar to the assumption in zero-shot setting\footnote{Note that our setting is
harder than zero-shot setting in \cite{LampertPAMI2013} as here the test set contains both
the known and novel objects} \cite{LampertPAMI2013}. We add the known novel words to the current
vocabulary and train the AE on the auxiliary text data. Once trained, we take the encoder weights
from the AE to initialize the question encoder in the VQA system. 
\vspace{0.3em} \\
\textbf{General setting.} 
Second, we assume that the novel words would be semantically similar to the known words and, thus,
expand the vocabulary by adding words, from the external dataset, which are within a
certain distance to the known words. The semantic word distance we use is the cosine distance
between the \texttt{word2vec} embeddings \cite{mikolov2013distributed} of the known and novel words.
This is a more relaxed assumption compared to the oracle setting and we call this the general
setting.

In practice, however, we found that the direct AE training was noisy in this general setting as the
vocabulary size increased by nearly $4\times$. We found that the noise and instability of the
the training mainly came from the word embeddings, \ie the projection of the one-hot word
representations before being fed to the recurrent unit, in the AE. 
In order to train it more effectively, we thus use a pre-training technique for initializing the
word embeddings of the AE as follows. We first train the AE on the BookCorpus with
the VQA vocabulary. We then take the words which appear both in the VQA vocabulary and the original
trained \texttt{word2vec} vocabulary. Using these words, we obtain a projection to align the the
\texttt{word2vec} vector space with the currently learned word embedding space. Formally, denote the
\texttt{word2vec} embedding matrix as $A_{w}$ and the VQA word embedding matrix as $A_{v}$, then we
find a projection matrix $M$, using least squares, as
\vspace{-0.6em}
\begin{align} 
A_{w} M &= A_{v}, \textrm{\ \ i.e., \ } M = (A_{w}^{\top} A_{w})^{-1} A_{w}^{\top} A_{v}.
\vspace{-1.5em}
\end{align}
Once the alignment matrix $M$ is available, the words in the general setting's vocabulary, which are
not in the VQA vocabulary, are computed as
\vspace{-0.6em}
\begin{equation} 
\hat{A}_{v}(w) = A_{v}(w)M,~~w\in \V_{g} \backslash \V_{v},
\vspace{-0.6em}
\end{equation}
where $\V_{g}$ is the vocabulary set of the general setting and $\V_{v}$ is that of VQA vocabulary. We
term this as vocabulary expansion from the first vocabulary to the second, similar in spirit to the
work by Kiros \etal \cite{kiros2015skip}. Going a step further, we initialize the word embedding
matrix in the AE using the estimated word vectors and train the AE again on BookCorpus, which we
finally use with the base architecture as in the other AE's above.
\vspace{0.3em} \\
\textbf{Auxiliary text as well as image data.}
Apart from using text only data to induce novel objects, we also attempt to use freely available
auxiliary image classification data, \eg ImageNet \cite{ILSVRC15}. The general philosophy stays the
same, we wish to train auto-encoders with the auxiliary data, but in this case such AE takes
multimodal input in the form of both text sentences and images, and decodes them back to the
sentences. We hope that such an AE\footnote{It is not strictly an AE as it is only decoding back the
text part and not the image part. We refrained from decoding back the images, as initial results
were not encouraging; also, image generation from encoded vectors is a complete challenging problem
in itself \cite{GoodfellowNIPS2014}.} will help induce novel
objects. To do so, we require paired image-text data and we use the two auxiliary datasets to
generate such paired data synthetically and weakly as follows.  We take images of the words
corresponding to objects in our text vocabulary from the classification dataset such as
ImageNet~\cite{ILSVRC15} and pair them with general sentences about the object from the text
dataset, \eg BookCorpus \cite{moviebook} or Wikipedia. Note that this is expected to be a noisy
paired data; we evaluate if such noise is tolerated by the AE to still give some improvement on the
VQA task by learning lexico-visual associations for novel objects.

Since the question encoder for the first architecture does not use the image as an input, we design
the corresponding multimodal AE as shown in Figure~\ref{figAE} (left). We take the output of the
multiply layer and use it to initialize the hidden state of the decoder. To keep the architecture
consistent with the text AE, we introduce a skip connection which feeds the final hidden state of
the encoder to the decoder's initial state. Adding such a skip layer ensures that the AE will use the
image encoding only if it is beneficial and we hope that this will add resilience to the noise in
the synthetically generated paired data. We, thus, effectively sum the final hidden state of the
encoder and the output of the multiply layer to obtain the initial decoder state. In case of the
second architecture, we just use the image encoding as the first input to the LSTM based AE, as
shown in the dashed part of Figure~\ref{figAE} (right).
\vspace{-0.7em}
\section{Experiments}
\vspace{-0.4em}
We now describe the experiments we performed to validate the method and study VQA when novel objects
are present in the test set. We first describe the datasets we used, followed by the new split we
created to have novel objects in the test set. We then give our quantitative and qualitative
results, with discussions.
\vspace{0.3em}\\
\textbf{VQA dataset} \cite{AntolICCV2015} is a publicly available benchmark which consists of images
obtained from the MSCOCO dataset~\cite{LinECCV14coco} and an abstract scenes dataset. The statistics
of the dataset are shown in Table~\ref{table:vqa_stats}. The models are evaluated on the VQA dataset
using the accuracy metric defined as
\begin{equation}
\text{\small acc} = \text{\small min}\bigg(\frac{\text{\small \# humans that provided that
answer}}{3}, 1\bigg).
\end{equation}
\begin{table}
	\centering
	\begin{tabular}{cc}
	\begin{minipage}{.45\linewidth}
	\scalebox{0.75}{
	\begin{tabular}{|l|c|}
		\hline
		\multicolumn{2}{|c|}{VQA dataset} \\
		\hline
		$\#$images & 204,721 \\
		$\#$ques& 614,163 \\
		$\#$ans per ques & 10 \\
		$\#$ques Types & more than 20 \\
		$\#$words per ans & one or more \\
		\hline
	\end{tabular}
	}	
	\end{minipage} \hfill
	\begin{minipage}{0.5\linewidth}
	\scalebox{0.75}{
		\begin{tabular}{|p{30mm}|c|}
		\hline
		\multicolumn{2}{|c|}{BookCorpus} \\	
		\hline
		$\#$books & 11,038 \\
		$\#$sentences & 74,004,228 \\
		$\#$unique words & 984,846,357 \\
		avg $\#$words / sent.& 13 \\
		\hline
	\end{tabular}
	}
	\end{minipage}
	\end{tabular}
	\vspace{-0.4em}
	\caption{Statistics of datasets used}
	\label{table:vqa_stats}
	\vspace{-1.5em}
\end{table}

\vspace{-1.3em}
\noindent
\textbf{BookCorpus} \cite{moviebook} dataset has text extracted from $11,038$ books available on the
web. Summary statistics of the dataset are shown in Table ~\ref{table:vqa_stats}. We created a split
consisting of $73,874,228$ training, $30,000$ validation and $100,000$ test sentences to train the AEs. 
\vspace{0.3em}\\
\textbf{ImageNet} dataset from the ILSVRC challenge~\cite{ILSVRC15} consists of images collected
from Flickr and other search engines. Each image is labelled with the presence or absence of one out
of $1000$ object categories. The training set consists of $1.2$ million training images, $50,000$
validation images and $100,000$ test images. We have used ImageNet to obtain images for the known
and unknown objects. 
\vspace{0.3em}\\
\textbf{Wikipedia.}
The text data obtained from BookCorpus did not have sentences containing some of the novel objects.
Also, the data obtained from BookCorpus was story oriented and not factual data, hence the sentences
containing certain objects did not describe the objects themselves, but just contained the objects
as a part of a narrative. In order to complement the data from BookCorpus and obtain descriptive
information about novel objects, we queried Wikipedia\footnote{Source:
\url{https://dumps.wikimedia.org/enwiki/latest/enwiki-latest-pages-articles.xml.bz2}} by searching for sentences containing the novel objects.   
\vspace{0.3em}\\
\textbf{Weak paired training data.}
To generate synthetic paired data, we consider all the objects from the oracle/general vocabulary
and find an intersection with the ImageNet classes. For each of the objects, we obtain $m$ random
images from the matched classes and $n$ random sentences containing the object from BookCorpus and
pair them to obtain $mn$ sets of paired images and sentences. In our case, we selected $m=20$ and
$n=20$. This constitutes the weak paired training data which amounted to approximately $ 0.25$
million samples for the oracle case and  $0.45$ million samples for the general case. 

\subsection{Proposed Novel Split for VQA dataset}
\label{labelSplit}
\vspace{-0.5em}
We create a new split of the VQA dataset to study the setting of novel objects at test time. We
obtain the train and validation split of the real scenes part of VQA dataset~\cite{AntolICCV2015}
and call this the original split. The questions from the train split are used for training and the
questions from the validation split are used for testing. Next, we divide the full set of images,
train and validation combined, into new train and test split as follows. For each of the questions
in the VQA dataset, we identify the nouns\footnote{We used NLTK's PerceptronTagger for obtaining
the nouns \url{http://www.nltk.org/_modules/nltk/tag/perceptron.html}} and create
a histogram of the types of questions each noun occurs in. We use normalized histograms to
cluster the nouns into $14$ clusters. We select $80\%$ of the nouns as known and $20\%$ of the nouns
as novel, randomly from each of the $14$ clusters. A question in the VQA dataset belongs to the new
test set if and only if at least one of the novel nouns occur in it. We randomly sample 5000
questions from the train split to create the validation split. The statistics of the original split
and the new proposed are shown in Table~\ref{tabSplitStats}---note that, while the original test split
contains $3178$ known objects out of $3330$ total, the proposed test split has only $2216$ known objects
out of a total of $3027$, \ie $811$ objects that appear in the test split were never seen (visually
or textually) in the VQA training data~\footnote{The design of the dataset leads to sharing of
images between the train and test splits; see supplementary material for detailed discussion.}. 

Further, Table~\ref{table:question_stats} shows the number of questions where $0$ to $5$ known objects
appear as well (in addition to at least one unknown object). We see that a large number of question,
\ie $32452$ contain only novel objects.

\begin{table}
	\centering
	\scalebox{0.85}{
	\begin{tabular}{|c|c|c|c||c|c|c|}
		\hhline{~|-|-|-|-|-|-|}
		\multicolumn{1}{c|}{} & \multicolumn{3}{c||}{$\#$ Questions} & \multicolumn{3}{c|}{$\#$ Objects}\\
		\hline
		Split & Train & Val & Test & Train & Test & Both \\ \hline
		Orig & 215375 & 0 & 121509 & 3625 & 3330 & 3178 \\
		Prop & 224704 & 5000 & 116323 & 2951 & 3027 & 2216 \\
		\hline 
	\end{tabular}
	}
	\vspace{-0.5em}
    \caption{Statistics of the dataset splits. 
    The proportion of seen test objects is $95.4\%$ in original \vs $73.2\%$ in proposed.
    } 
	\vspace{-1em}
	\label{tabSplitStats}
\end{table}

\begin{table}
	\centering
	\scalebox{0.85}{
	\begin{tabular}{|c||c|c|c|c|c|c|}
		\hline
		$\#$Known objs& 0 & 1 & 2 & 3 & 4 & 5 \\
		\hline
		$\#$Questions  & 32452 & 35300 & 12593 & 2605 & 501 & 48 \\
		\hline
	\end{tabular}
	}
    \vspace{-0.5em}
	\caption{
    The number of questions with specific number of known words in test set. 
    } 
    \vspace{-1em}
    \label{table:question_stats}
\end{table}

\begin{table}
\centering
\scalebox{0.73}{
\begin{tabular}{|c|c|c|c|c||c|c|c|c|}
    \multicolumn{1}{c}{ } & \multicolumn{8}{c}{Architecture 1} \\
	\cline{2-9}
	\multicolumn{1}{c|}{ } & \multicolumn{4}{c||}{Open Ended Questions} 
                          & \multicolumn{4}{c|}{Multiple Choice Questions} \\
	\hline 
	Split & Ov.all & Oth. & Num. & Y/N & Ov.all & Oth. & Num. & Y/N   \\
	\hline
	Orig& 54.23 & 40.34 & 33.27 & 79.82 & 59.30 & 50.16 & 34.41 & 79.86 \\
	Novel & 39.38 & 23.07 & 27.52 & 74.02 & 46.54 & 34.91 & 29.39 & 74.10 \\
	Drop  & \textbf{14.85} & \textbf{17.27} & \textbf{5.75} & \textbf{5.8} & \textbf{12.76} & \textbf{15.25} & \textbf{5.02} & \textbf{5.76} \\
	\hline
    \multicolumn{1}{c}{ } & \multicolumn{8}{c}{Architecture 2} \\
	\cline{2-9}
	\multicolumn{1}{c|}{ } & \multicolumn{4}{c||}{Open Ended Questions} 
                          & \multicolumn{4}{c|}{Multiple Choice Questions} \\
	\hline
	Split & Ov.all & Oth. & Num. & Y/N & Ov.all & Oth. & Num. & Y/N   \\
	\hline
	Orig& 48.75 & 33.31 & 31.42 & 74.20 & 54.94 & 45.24 & 32.95 & 75.28 \\
	Novel & 34.97 & 16.98 & 28.27 & 71.06  & 42.83 & 30.16 & 29.42 & 71.12 \\
	Drop  & \textbf{13.78} & \textbf{16.33} & \textbf{3.15} & \textbf{3.14} & \textbf{12.11} & \textbf{15.08} & \textbf{3.53} & \textbf{4.16} \\
	\hline
\end{tabular}
}
\vspace{-0.5em}
\caption{
The drop in performance for novel word setting.
}
\vspace{-1em}
\label{table:baseline_results}
\end{table}

\vspace{0.3em}
\noindent 
\textbf{Implementation details.}
In the case of Architecture 1, we used the default settings of $200$ dimensional word encoding size,
$512$ RNN hidden layer size and $2$ RNN layers for computing the results on the case of training
only with VQA dataset. To avoid very long training times, with the above large parameter values for
other architectures, we selected $512$ dimensional word encoding, $512$ RNN hidden layer size and
$1$ RNN layer for computing all our results. We observed that this did not affect our results
appreciably.  Similarly, in Architecture 2, we used $512$ dimensional word encoding, $512$ RNN
hidden layer size and $1$ RNN layer throughout all our experiments.
\vspace{-0.3em}
\subsection{Quantitative Results}
\begin{figure*}
\centering
\includegraphics[width=0.95\linewidth, trim=0 170 50 5, clip]{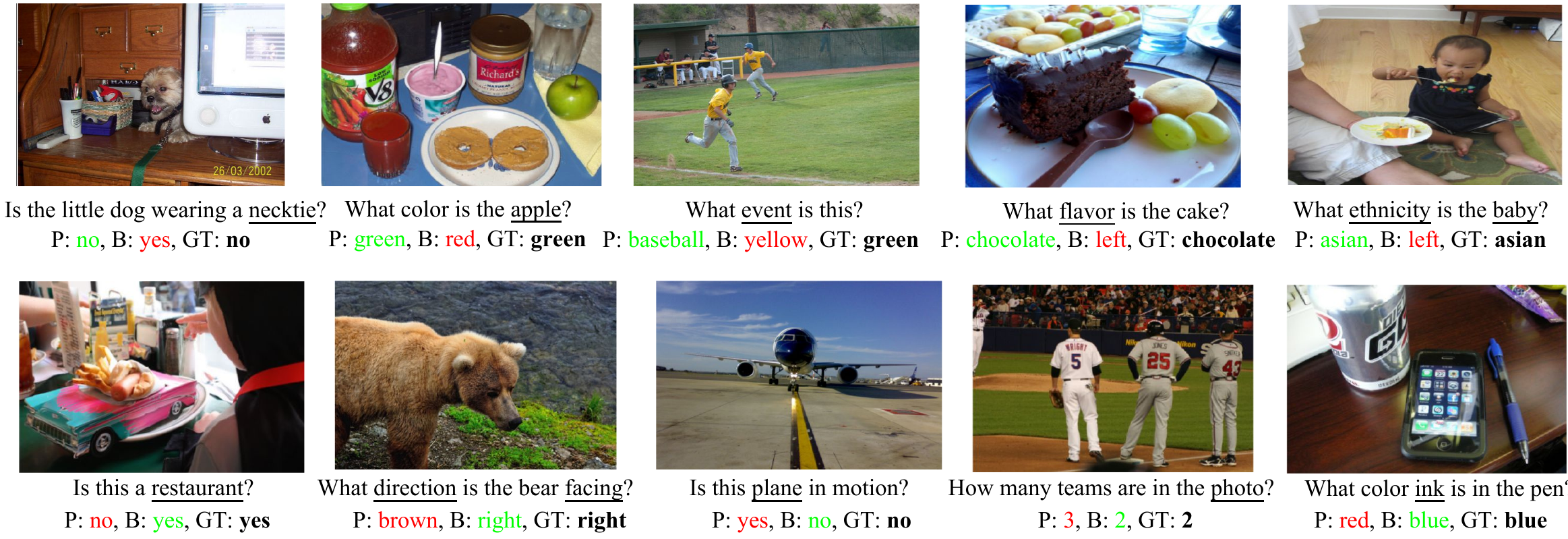}
\vspace{-0.5em}
\caption{Qualitative examples highlighting the success and failure cases of our proposed
model(P)~\cf the baseline model(B) and the ground truth(GT). The novel concepts are underlined in
the question.} \label{figQual}
\vspace{-0.2em}
\end{figure*}

\vspace{-0.3em}
\label{secQuant}
Our overall results for the two architectures are shown in Table~\ref{tabComparison}. The results
are split into the standard question types of Overall, Others, Numbers and Yes/No. We also introduce
the Novel question type which consists of all the questions which contain only the novel objects and
no known objects ($32452$ questions from Table~\ref{table:question_stats}). This helps us analyse the
performance of novel object VQA without interference from the known objects. The image
feature, auxilliary data and the vocabulary used for each of the experiments has been specified. The
image feature can be \texttt{VGG}, \texttt{INC} (Inception), \texttt{EF} (Early fusion of VGG, INC)
or \texttt{LF} (Late fusion of VGG, INC), the auxilliary data can be \texttt{none} (baseline),
\texttt{text} (BookCorpus pre-trained AE) or \texttt{text+im} (BookCorpus + WeakPaired data
pre-trained AE) and the vocabulary can be \texttt{train} (only words from train data of novel
split), \texttt{oracle} (oracle case), \texttt{gen} (general case) or \texttt{gen(exp)} (vocabulary
expansion in general case).  We analyse our results in terms of the need to incorporate novel words,
effects of different features, vocabulary expansion and pre-training methods on the overall
performance. In the following, we refer to a cell in the tables with the Architecture number, the
row number and the type of questions (others, numbers \etc in Open Ended or Multiple Choice
questions). If we do not specify the sub-type of questions for OEQ or MCQ, then we are discussing
the overall averages for these two types.
\vspace{0.3em} \\
\noindent
\textbf{Performance on original \vs novel split.}
Table~\ref{table:baseline_results} gives the results of the two architectures 
on the original and novel splits, respectively, without using any data outside of the VQA
dataset\footnote{While the training/testing data are not same, and hence the performances are not
directly comparable, we note that the amount of training data is $\sim 4\%$ more for the models
trained in the novel setting (\autoref{tabSplitStats}). If the difficulties of the settings were
similar, the novel models should have, arguably, done better due to access to more training data.}.
We obseve a severe drop in perfomance, \eg Architecture 1 (2) drops by $27\%\ (28\%)$ on average for
the open ended questions, and $21\%\ (22\%)$ on the multiple choice ones. This highlights the fact
that the current methods are not capable of generalizing on VQA to novel objects when not
explicitly trained to do so. This empirically verifies the argument that VQA in the novel object
setting is a challenging problem and deserves attention on its own.
\vspace{0.3em} \\
\textbf{Na\"ive pre-training is not sufficient.}
An obvious first argument, as discussed in Sec.~\ref{secNovObj}, is that pre-training the text model
on the large amount of auxiliary text data, might make it better and hence lead to improved
performance, even when the vocabulary is kept the same as the original one (which does not contain
the novel words). We tested this hypothesis and found it to not be true.  While text only
pre-training (rows A1.b, A2.b in Table~\ref{tabComparison}) provided some improvements over the
baseline (row A1.a) in most cases, \eg $39.38$ to $40.09$ in A1.b OEQ, $46.47$ to
$47.01$ in A1.f MCQ, they were generally minor, especially in the high performing models; some
isolated larger improvements did happen, \eg $+6.6\%$ ($34.97$ to $37.30$) in A2.b OEQ, but they were not
consistent and happened in the relatively low performing cases only. However, the text only
pre-trained models \emph{with} the oracle and general vocabularies provided consistent improvements,
\eg $+2.7\%$ ($39.38$ to $40.44$) in A1.c OEQ, $+2.3\%$ ($40.27$ to $41.19$) A1.g OEQ, $+7.7\%$ ($34.97$ to
$37.68$) in A2.c OEQ and $+2.3\%$ ($37.66$ to $38.53$) in A2.g OEQ, as they were capable of
understanding novel objects. Hence, we conclude that simple pre-training without adding the novel
objects to the vocabulary is not sufficient for novel object test setting in VQA. 
\vspace{0.8em} \\
In the following, all the discussion are \wrt methods using vocabularies incorporating novel objects.
\vspace{0.3em} \\
\textbf{Comparison of architectures.}
We found that that Architecture 1 generally performed better than Architecture 2, \eg $39.38$ on
A1.a OEQ \vs $34.97$ on A2.a OEQ, $39.56$ in A1.k OEQ \vs $35.65$ on A2.k OEQ. The relative
improvements obtained with the better performing architecture over the corresponding baseline were,
unfortunately, generally lesser, \eg $+6.2\%$ and $+4.2\%$ in A1.i OEQ and MCQ \vs $+11.1\%$ and
$+8.5\%$ in A2.i OEQ and MCQ, both with early fusion of VGG and Inception features, respectively,
indicating that it is more difficult to improve performance for more saturated methods. We do, however, see 
consistent improvements in majority of cases for both the Architectures, supporting the proposed
method.
\vspace{0.3em} \\
\textbf{Auxiliary text data.}
The models initialized from auxiliary text data, with both oracle and general vocabularies, provide
significant improvement in the Yes/No, \eg $+5.6\%$ ($71.06$ to $75.06$) in A2.c OEQ, $+4.4\%$
 ($73.25$ to $76.49$) in A1.q OEQ, and Novel questions, such as $2.5\%$ ($48.03$ to $49.23$) in A1.g
OEQ, $+5.2\%$ ($44.60$ to $46.93$) in A2.c OEQ. The proposed model improves on Yes/No
questions as they generally have a central
object, \eg `is the little dog wearing a necktie?' (Fig.~\ref{figQual}, image on left-top), and when this
object (necktie here) is unknown the baseline model fails to understand the question. Similar trend is
visible in the `Novel' type. 

The effect of the general (automatic) vocabulary expansion technique is similar to the oracle case,
where the novel objects are assumed to be known \emph{a priori}. The overall results with oracle
vocabulary \vs general vocabulary are similar, \ie $41.84$ \vs $41.82$, $48.87$ \vs $48.35$, for
A1.(i,s) OEQ and MCQ, respectively, and $39.49$ \vs $39.91$ and $46.40$ \vs $46.99$ for A2.(i,t) OEQ
and MCQ, respectively. Thus, we conclude that the proposed method is capable of leveraging auxiliary
text data to improve novel object VQA, in the automatic setting when the minimal assumption is
made that the novel words are expected to be semantically similar to the known words.
\vspace{0.2em} \\
\textbf{Vocabulary expansion.}
Generally, the accuracy of the system improves with vocabulary expansion on the Yes/No and Novel
question categories when compared to the accuracy of the non-expanded setting, \eg $75.48, 48.78$ in
A1.p OEQ \vs $76.49, 49.36$ in A1.q OEQ and $74.38, 51.29$ in A2.l MCQ \vs $75.28, 52.47$ in A2.m
MCQ. This follows the trend from auxiliary text data where we observed similar improvements, and is
expected since vocabulary expansion is simply a better way to perform text only pre-training. 
\vspace{0.2em} \\
\textbf{Auxiliary text and image data.}
Using both auxiliary datasets of text and image, as proposed, led to consistent but small
improvements over using only auxiliary text datasets. As an example, consider Inception features for
Architecture 2 in A2.f--h OEQ. The baseline of $37.66$ is improved to $38.53$ ($+2.3\%$) by oracle
vocabulary expansion and use of auxiliary text data which is further improved to $38.75$ ($+2.9\%$)
when using both auxiliary data of both text and image---the major improvement comes from using text
data and a further small improvement is achieved by using image data as well. We believe that since
the text data is relatively clean and rich, it provides good semantic ground for the model to
understand the novel objects, while the noisy method of generating weak text-image paired data as
proposed is not able to supplement it significantly, and sometimes even deteriorates it slightly.
Also, since the image model may have seen the novel objects \emph{a priori}, this may not have a
significant impact on the overall results.
\vspace{0.3em} \\
\textbf{Additional observations.}
Apart from the above main observations, we found that Inception features were generally better than
VGG features for VQA. However, most of the improvement of Inception over VGG features was in the
``others" category, \eg $23.07$ in A1.a OEQ \vs $24.54$ A1.e OEQ and $30.74$ in A2.k MCQ \vs $31.87$
in A2.o MCQ. The Inception baseline models do not generally perform better than VGG baseline models
on the ``Novel" questions, especially in Architecture 1 which is the stronger architecture.
Therefore, improving image features alone is not sufficient for better novel objects based VQA. This
is expected since the text model is still the same and without improvements in the text model or
better joint modelling, we cannot expect a significant difference in performance on novel objects.

\begin{table*}
\small
\centering
\scalebox{0.9}{
\def\arraystretch{.88}
\newcolumntype{C}{>{\centering\arraybackslash}p{3em}}
\begin{tabular}{|ccll|CCCCC||CCCCC|}
	\multicolumn{2}{c}{ }& \multicolumn{12}{c}{Architecture 1 (A1)} \\
	\cline{5-14}
	\multicolumn{3}{c}{ } & \multicolumn{1}{c|}{} & \multicolumn{5}{c||}{Open Ended Questions (OEQ)}
    & \multicolumn{5}{c|}{Multiple Choice Questions (MCQ)} \\
	\hline
	Row & Feat & Aux & Vocab &  Overall & Others & Numbers & Yes/No & Novel & Overall & Others & Numbers & Yes/No & Novel \\ 
    \hline
	\cellcolor{Gray}a & \cellcolor{Gray}VGG & \cellcolor{Gray}none & \cellcolor{Gray}oracle&  \cellcolor{Gray}39.38 & \cellcolor{Gray}23.07 & \cellcolor{Gray}27.52 & \cellcolor{Gray}74.02 & \cellcolor{Gray}47.56 & \cellcolor{Gray}46.54 & \cellcolor{Gray}34.91 & \cellcolor{Gray}29.39 & \cellcolor{Gray}74.10 & \cellcolor{Gray}52.32 \\
	b & VGG & text & train & 40.09 & 23.46 & 28.85 & 75.14 & 48.75 & 47.22 & 35.32 & 20.36 & 75.21 & 53.39 \\
	c & VGG & text & oracle& 40.44 & 23.42 & 28.24 & 76.52 & 48.95 & 47.65 & 35.39 & 29.89 & 76.60 & 53.77 \\
	d & VGG & text+im & oracle& 40.49 & 23.35 & 28.32 & 76.79 & 48.89 & 47.38 & 34.76 & 30.04 & 76.87 & 53.80 \\
	\hline
	\cellcolor{Gray}e & \cellcolor{Gray}INC & \cellcolor{Gray}none & \cellcolor{Gray}oracle & \cellcolor{Gray}40.27 & \cellcolor{Gray}24.54 & \cellcolor{Gray}28.02 & \cellcolor{Gray}73.95 & \cellcolor{Gray}48.03 & \cellcolor{Gray}46.47 & \cellcolor{Gray}34.84 & \cellcolor{Gray}29.41 & \cellcolor{Gray}74.00 & \cellcolor{Gray}52.19 \\
	f & INC & text & train & 40.18 & 24.12 & 28.25 & 74.37 & 48.10 & 47.01 & 35.43 & 29.91 & 74.46 & 52.80 \\
	g & INC & text & oracle& 41.19 & 24.98 & 28.44 & 75.93 & 49.23 & 47.87 & 36.00 & 30.24 & 76.04 & 53.88 \\
	h & INC & text+im & oracle & 40.73 & 24.12 & 27.80 & 76.03 & 48.61 & 47.23 & 34.99 & 29.58 & 76.12 & 53.18 \\
	\hline
	i & EF & text & oracle & 41.84 & 25.69 & 27.93 & 76.87 & 49.76 & 48.47 & 36.62 & 29.75 & 76.96 & 54.40 \\
	j & LF & text & oracle & 41.46 & 25.39 & 28.66 & 75.95 & 49.32 & 48.22 & 36.33 & 30.26 & 76.54 & 54.04 \\
	\hline
	\hline
	\cellcolor{Gray}k & \cellcolor{Gray}VGG & \cellcolor{Gray}none & \cellcolor{Gray}gen & \cellcolor{Gray}39.56 & \cellcolor{Gray}23.18 & \cellcolor{Gray}28.47 & \cellcolor{Gray}74.06 & \cellcolor{Gray}48.02 & \cellcolor{Gray}46.23 & \cellcolor{Gray}34.27 & \cellcolor{Gray}29.92 & \cellcolor{Gray}74.13 & \cellcolor{Gray}52.44 \\
	l & VGG & text & gen & 40.53 & 23.62 & 28.93 & 76.20 & 49.00 & 47.50 & 35.26 & 30.10 & 76.27 & 53.45 \\
	m & VGG & text & gen(exp) & 40.76 & 23.89 & 28.19 & 76.69 & 49.05 & 47.82 & 35.67 & 29.40 & 76.79 & 53.75 \\
	n & VGG & text+im & gen(exp) & 40.34 & 23.09 & 29.25 & 76.49 & 49.25 & 47.36 & 34.82 & 30.31 & 76.60 & 53.92 \\
	\hline
	\cellcolor{Gray}o & \cellcolor{Gray}INC & \cellcolor{Gray}none & \cellcolor{Gray}gen & \cellcolor{Gray}40.25 & \cellcolor{Gray}24.86 & \cellcolor{Gray}28.12 & \cellcolor{Gray}73.25 & \cellcolor{Gray}47.77 & \cellcolor{Gray}46.53 & \cellcolor{Gray}35.28 & \cellcolor{Gray}29.56 & \cellcolor{Gray}73.33 & \cellcolor{Gray}52.07 \\
	p & INC & text & gen & 40.76 & 24.54 & 28.14 & 75.48 & 48.78 & 46.87 & 34.77 & 28.95 & 75.56 & 52.83 \\
	q & INC & text & gen(exp) & 41.39 & 24.96 & 28.83 & 76.49 & 49.36 & 47.88 & 35.74 & 30.09 & 76.62 & 53.77 \\
	r & INC & text+im & gen(exp) & 40.42 & 23.77 & 27.98 & 75.88 & 48.77 & 46.87 & 34.53 & 29.11 & 75.99 & 52.87 \\
	\hline
	s & EF & text & gen(exp) & 41.82 & 25.72 & 28.51 & 76.55 & 49.60 & 48.35 & 36.57 & 29.81 & 76.65 & 53.92 \\
	t & LF & text & gen(exp) &  39.66 & 24.03 & 27.65 & 73.07 & 47.34 & 47.26 & 35.37 & 29.42 & 75.53 & 53.13 \\
	\hline
\end{tabular}
} 
\\
\scalebox{0.9}{
\def\arraystretch{.88}
\newcolumntype{C}{>{\centering\arraybackslash}p{3em}}
\begin{tabular}{|ccll|CCCCC||CCCCC|}
    \multicolumn{11}{c}{ } \\
	\multicolumn{2}{c}{ }& \multicolumn{12}{c}{Architecture 2 (A2)} \\
	\cline{5-14}
	\multicolumn{3}{c}{ } & \multicolumn{1}{c|}{} & \multicolumn{5}{c||}{Open Ended Questions (OEQ)}
    & \multicolumn{5}{c|}{Multiple Choice Questions (MCQ)} \\
    \hline
	Row & Feat & Aux & Vocab & Overall & Others & Numbers & Yes/No & Novel & Overall & Others & Numbers & Yes/No & Novel \\ 
    \hline
	\cellcolor{Gray}a & \cellcolor{Gray}VGG & \cellcolor{Gray}none & \cellcolor{Gray}oracle & \cellcolor{Gray}34.97 & \cellcolor{Gray}16.98 & \cellcolor{Gray}28.27 & \cellcolor{Gray}71.06 & \cellcolor{Gray}44.60 & \cellcolor{Gray}42.83 & \cellcolor{Gray}30.16 & \cellcolor{Gray}29.42 & \cellcolor{Gray}71.12 & \cellcolor{Gray}49.38 \\
	b & VGG & text & train & 37.30 & 19.50 & 26.24 & 74.48 & 45.71 & 44.30 & 31.26 & 27.09 & 74.55 & 50.31 \\
	c & VGG & text & oracle & 37.68 & 19.50 & 28.28 & 75.06 & 46.93 & 45.12 & 31.91 & 29.64 & 75.11 & 51.67 \\
	d & VGG & text+im & oracle & 38.06 & 20.15 & 28.45 & 74.98 & 47.54 & 45.80 & 32.96 & 30.30 & 75.10 & 52.66 \\
    \hline
	\cellcolor{Gray}e & \cellcolor{Gray}INC & \cellcolor{Gray}none & \cellcolor{Gray}oracle & \cellcolor{Gray}37.66 & \cellcolor{Gray}20.18 & \cellcolor{Gray}28.32 & \cellcolor{Gray}73.69 & \cellcolor{Gray}46.50 & \cellcolor{Gray}44.59 & \cellcolor{Gray}31.77 & \cellcolor{Gray}29.32 & \cellcolor{Gray}73.77 & \cellcolor{Gray}50.98 \\
	f & INC & text & train & 37.37 & 20.00 & 25.90 & 73.89 & 45.54 & 44.40 & 31.83 & 26.59 & 73.96 & 50.27 \\
	g & INC & text & oracle & 38.53 & 20.79 & 28.07 & 75.39 & 47.55 & 45.85 & 32.98 & 29.37 & 75.49 & 52.32 \\
	h & INC & text+im & oracle & 38.75 & 21.12 & 28.96 & 75.20 & 47.95 & 46.07 & 33.32 & 30.13 & 75.34 & 52.53 \\
    \hline
	i & EF & text & oracle & 38.85 & 21.18 & 28.43 & 75.57 & 48.00 & 46.47 & 33.76 & 30.58 & 75.66 & 53.15 \\
	j & LF & text & oracle & 39.49 & 22.02 & 28.71 & 75.95 & 48.47 & 46.40 & 33.56 & 29.56 & 76.04 & 52.86 \\
    \hline
    \hline
	\cellcolor{Gray}k & \cellcolor{Gray}VGG & \cellcolor{Gray}none & \cellcolor{Gray}gen & \cellcolor{Gray}35.65 & \cellcolor{Gray}17.33 & \cellcolor{Gray}26.62 & \cellcolor{Gray}73.14 & \cellcolor{Gray}45.19 & \cellcolor{Gray}43.64 & \cellcolor{Gray}30.74 & \cellcolor{Gray}27.40 & \cellcolor{Gray}73.28 & \cellcolor{Gray}50.29 \\
	l & VGG & text & gen & 37.66 & 19.95 & 27.73 & 74.31 & 46.64 & 44.99 & 32.19 & 29.01 & 74.38 & 51.29 \\
	m & VGG & text & gen(exp) & 38.00 & 20.21 & 26.77 & 75.21 & 46.84 & 45.96 & 33.32 & 29.26 & 75.28 & 52.47 \\
	n & VGG & text+im & gen(exp) & 37.92 & 20.21 & 27.90 & 74.59 & 45.58 & 45.58 & 33.04 & 28.99 & 74.67 & 52.15 \\
    \hline
	\cellcolor{Gray}o & \cellcolor{Gray}INC &\cellcolor{Gray}none & \cellcolor{Gray}gen & \cellcolor{Gray}37.29 & \cellcolor{Gray}19.59 & \cellcolor{Gray}28.76 & \cellcolor{Gray}73.50 & \cellcolor{Gray}46.16 & \cellcolor{Gray}44.63 & \cellcolor{Gray}31.87 & \cellcolor{Gray}27.40 & \cellcolor{Gray}73.28 & \cellcolor{Gray}50.29 \\
	p & INC & text & gen & 38.23 & 20.89 & 28.11 & 74.22 & 46.94 & 45.23 & 32.54 & 29.55 & 74.31 & 51.67 \\
	q & INC & text & gen(exp) & 37.99 & 20.59 & 26.30 & 74.65 & 46.31 & 45.89 & 33.54 & 29.01 & 74.72 & 51.84 \\
	r & INC & text+im & gen(exp) & 38.20 & 20.49 & 27.79 & 75.00 & 46.97 & 45.65 & 32.94 & 28.96 & 75.08 & 51.93 \\
    \hline
	s & EF & text & gen(exp) & 38.37 & 21.13 & 28.82 & 74.00 & 47.10 & 45.46 & 33.00 & 29.99 & 74.05 & 51.89\\
	t & LF & text & gen(exp) & 39.91 & 22.75 & 28.90 & 75.87 & 48.48 & 46.99 & 34.55 & 30.24 & 75.94 & 53.26\\
    \hline
\end{tabular}
} 
\vspace{-1em}
\caption{Perfomances of the different models in the novel object setting for VQA}
\vspace{-1em}
\label{tabComparison}
\end{table*}

\vspace{-0.3em}
\subsection{Qualitative results}
\vspace{-0.5em}
\label{expQual}
Figure~\ref{figQual} shows some example images with the questions and their answers from the
different methods. In the first row, we can observe that proposed model (corresponding to A1.s) has
successfully induced the concepts of mouse, apple, event and flavor into the VQA framework whereas the
baseline (corresponding to A1.k) has failed to reason based on them. Some of the failure cases of
the proposed model are illustrated in the second row. It has failed to induce the concepts of
restaurant and direction. We also feel that in the last 2 cases, it has predicted purely based on
the text modelling. For example, in the fourth case it says that the plane is in motion and in the
fifth case it says the ink is red. This could be because it witnessed similar textual examples and
the image is not convincing enough for it to say otherwise. 
\vspace{-0.20cm}
\section{Conclusion}
\vspace{-0.5em}
We presented a new task of VQA based on novel objects which were not seen during training. This is a
relevant setting as in real world, owing to the heavy tailed distribution of the visual
categories, many rare objects are not expected to have annotations. We showed that this is a
challenging scenario and directly testing the models which had not seen the objects during training
leads to substantial degradation in performances of up to $28\%$. We proposed to use auxiliary
datasets of text, \eg books and Wikipedia, and images, \eg ImageNet, to make the system aware of the
novel objects it might encounter during testing. We showed that increasing the vocabulary, to
include possible novel words, is important and a simple pre-training on the auxiliary data is not
sufficient. We proposed two methods for incorporating novel objects in VQA systems. 
In the first oracle method, we assumed that the novel objects that would appear are given to us,
while in the second we made the weaker assumption that the novel words will be semantically similar
to the known words. We also proposed a method to use external labeled image datasets to form noisy
image-text pairs for pre-training the VQA architectures. Our results demonstrated that making the
model aware of novel objects using vocabulary expansion and pre-training on external text datasets
significantly improves the performance for VQA in novel test object setting \eg by $+3.4\%$ on the
Yes/No questions, $+3.6\%$ on Numbers, $+11.48\%$ on Others and $+4.8\%$ on Novel for Architecture 1
and by $+6.76\%$ on the Yes/No questions, $+2.2\%$ on Numbers, $+24.4\%$ on Others and $+8.7\%$ on
Novel for Architecture $2$ in OpenEnded questions category. However, the gains from external image
datasets were either absent or were only modest. We believe that the external text datasets provided
a clean and rich source of knowledge while the paired image information was noisy and hence
relatively less effective. 
{\noindent
\subsection*{Acknowledgements} 
\vspace{-0.3em}
GS acknowledges support via the Early Career Research Award from SERB India (File \#
ECR/2016/001158) and the Research-I foundation at IIT Kanpur.
}

{\small
\bibliographystyle{ieee}
\bibliography{novel_object_cvpr2017.bib}
}

\appendix
\section*{SUPPLEMENTARY MATERIAL}
\subsection*{Possible extensions}
Here, we tackle the problem (P1) of answering \emph{known} (e.g.\ similar to those in train set)
questions containing \emph{novel} objects and having \emph{known} answers, at test time. More challenging cases include (P2) answering novel
questions about novel objects (as suggested by AR2) and (P3) generating answers containing novel objects
(as suggested by AR5). While problems P2 and P3 are more difficult problems than P1,
we highlight that P1 is itself a very challenging subproblem which has not been addressed so far. In the proposed setting, questions that come under
P1 account for a significant fraction of the test questions ($71.79\%$, $83,508$ out of $116,323$). If a
perfect model were available for P1, then the overall accuracy would
be $71.79\%$.  However, current methods obtain accuracies of around $40\%$, highlighting that P1
itself is very hard and arguably should be the first stepping stone in this direction.

\begin{table*}[t]
\centering
    \begin{tabular}{ |c|c|c|c|c|c|c|c|c|c|c|c|c|c| } 
    \cline{7-14}
    \multicolumn{6}{c|}{ } & \multicolumn{4}{c|}{Open Ended Questions} 
                  & \multicolumn{4}{c|}{Multiple Choice Questions} \\

    \hline
    arch & feat & model & aux & vocab & set & ov & oth & num & y/n & ov & oth & num & y/n\\ 
    \hline
    \parbox[c]{2mm}{\multirow{12}{*}{\rotatebox[origin=c]{90}{Architecture 1}}}
    & \parbox[c]{2mm}{\multirow{6}{*}{\rotatebox[origin=c]{90}{VGG}}} &
    \multirow{2}{*}{1} & \multirow{2}{*}{none} & \multirow{2}{*}{oracle} & s1 & 39.64 & 21.4 & 28.94 & 73.22 & 46.39 & 33.14 & 30.64 & 73.27 \\ 
    & & & & & s2 & 39.33 & 23.37 & 27.24 & 74.19 & 46.57 & 35.23 & 29.15 & 74.27 \\
    & & \multirow{2}{*}{2} & \multirow{2}{*}{text} & \multirow{2}{*}{train} & s1 & 40.39 & 21.82 & 29.49 & 74.61 & 47.20 & 33.66 & 31.12 & 74.68 \\
    & & & & & s2 & 40.04 & 23.75 & 28.73 & 75.25 & 47.23 & 35.62 & 30.21 & 75.33\\
    & & \multirow{2}{*}{3} & \multirow{2}{*}{text} & \multirow{2}{*}{oracle} & s1 & 40.89 & 21.91 & 28.81 & 76.15 & 47.91 & 34.10 & 30.66 & 76.21 \\
    & & & & & s2 & 40.35 & 23.69 & 28.13 & 76.60 & 47.60 & 35.62 & 29.73 & 76.68 \\
    \cline{2-14}
    & \parbox[c]{2mm}{\multirow{6}{*}{\rotatebox[origin=c]{90}{INCEP}}}
    & \multirow{2}{*}{4} & \multirow{2}{*}{none} & \multirow{2}{*}{oracle} & s1 & 40.71 & 23.00 & 29.88 & 73.47 & 46.72 & 33.52 & 31.01 & 73.52 \\
    & & & & & s2 & 40.19 & 24.82 & 27.65 & 74.04 & 46.42 & 35.07 & 29.09 & 74.11 \\
    & & \multirow{2}{*}{5} & \multirow{2}{*}{text} & \multirow{2}{*}{train} & s1 & 40.40 & 22.61 & 28.77 & 73.53 & 46.79 & 33.70 & 30.47 & 73.61 \\
    & & & & & s2 & 40.14 & 24.39 & 28.15 & 74.55 & 47.05 & 35.74 & 29.80 & 74.64 \\
    & & \multirow{2}{*}{6} & \multirow{2}{*}{text} & \multirow{2}{*}{oracle} & s1 & 41.20 & 23.20 & 28.61 & 74.99 & 47.76 & 34.63 & 29.97 & 75.11 \\
    & & & & & s2 & 41.19 & 25.30 & 28.41 & 76.12 & 47.89 & 36.24 & 30.29 & 76.23 \\
    \hline
    \hline
    \parbox[c]{2mm}{\multirow{12}{*}{\rotatebox[origin=c]{90}{Architecture 2}}} &
    \parbox[c]{2mm}{\multirow{6}{*}{\rotatebox[origin=c]{90}{VGG}}}
    & \multirow{2}{*}{1} & \multirow{2}{*}{none} & \multirow{2}{*}{oracle} & s1 & 35.45 & 15.53 & 28.86 & 70.55 & 43.02 & 28.90 & 29.79 & 70.57 \\ 
    & & & & & s2 & 34.87 & 17.24 & 28.16 & 71.17 & 42.80 & 30.39 & 29.35 & 71.24 \\
    & & \multirow{2}{*}{2} & \multirow{2}{*}{text} & \multirow{2}{*}{train} & s1 & 37.52 & 17.37 & 27.37 & 74.12 & 44.45 & 29.63 & 27.85 & 74.20 \\
    & & & & & s2& 37.26 & 19.88 & 26.02 & 74.56 & 44.27 & 31.54 & 26.95 & 74.62 \\
    & & \multirow{2}{*}{3} & \multirow{2}{*}{text} & \multirow{2}{*}{oracle} & s1 & 37.99 & 17.66 & 28.16 & 74.78 & 45.04 & 30.00 & 29.47 & 74.83 \\
    & & & & & s2 & 37.62 & 19.83 & 28.30 & 75.11 & 45.13 & 32.25 & 29.67 & 75.17 \\
    \cline{2-14}
    & \parbox[c]{2mm}{\multirow{6}{*}{\rotatebox[origin=c]{90}{INCEP}}}
    & \multirow{2}{*}{4} & \multirow{2}{*}{none} & \multirow{2}{*}{oracle} & s1 & 37.95 & 18.50 & 28.64 & 73.10 & 44.59 & 30.20 & 29.52 & 73.17 \\
    & & & & & s2 & 37.61 & 20.49 & 28.25 & 73.81 & 44.59 & 32.05 & 29.28 & 73.89 \\
    & & \multirow{2}{*}{5} & \multirow{2}{*}{text} & \multirow{2}{*}{train} & s1 & 37.73 & 17.98 & 27.63 & 73.64 & 44.48 & 29.93 & 28.07 & 73.73 \\
    & & & & & s2 & 37.30 & 20.36 & 25.56 & 73.94 & 44.39 & 32.17 & 26.30 & 74.00 \\
    & & \multirow{2}{*}{6} & \multirow{2}{*}{text} & \multirow{2}{*}{oracle} & s1 & 38.71 & 18.87 & 28.26 & 74.87 & 45.80 & 31.31 & 29.44 & 74.95 \\
    & & & & & s2 & 38.49 & 21.14 & 28.04 & 75.50 & 45.85 & 33.28 & 29.35 & 75.60\\

    \hline
    \end{tabular}
    \vspace{1em}
    \caption{Arch.\ 1 (top) and Arch.\ 2 (bottom) with VGG and INCEP features: s1 is images exclusive to test and s2 is common images between train and test; in both cases questions are test only.}
    \label{tab:overfit}
    \vspace{-1em}
\end{table*}

\subsection*{Image sharing}
Image sharing takes place in our proposed split, statistics  are shown in Tab.~\ref{tab:sharing}.
We claim that overfitting does not happen and justify the claim with the performances of the system
computed separately on the common and exclusive parts of test set; Tab.~\ref{tab:overfit} gives
these performances and we see that the differences in overall performance are very small ($\leq 0.5$) in all cases.
Also, the improvements obtained by various models over the baseline model are similar 
in the common and exclusive parts of the test set.

We would like to also highlight that sharing images does not make the task easier or the split prone
to overfitting (as already demonstrated by the results above). Even if the same image is present in
train and test set, the object being queried for is different at train and test time (by design of
the split). Hence, the system can not memorize or overfit on the train set \emph{and} give good
performance on the test set. 

\begin{table*}
    \centering
    \begin{tabular}{ |p{1.2cm}|p{1cm}|p{1cm}|c|c| } 
        \hline
        \multirow{2}{*}{$\#$ Images} & \multicolumn{2}{|c|}{Common to train and test}  & Train only & Test only \\ 
        \cline{2-5}
        & \multicolumn{2}{|c|}{73487} & 43583 & 6216 \\ 
        \hline
        \multirow{2}{*}{\parbox[t]{2cm}{$\#$ Corres. \\ Questions}} & Train & Test & Train & Test \\
        \cline{2-5}
        & 108857 & 97675 & 115847 & 18648 \\ 
        \hline
    \end{tabular}
    \vspace{1em}
	\caption{Statistics of images in train/test splits} 
    \label{tab:sharing}
\vspace{-1em}
\end{table*}

\end{document}